\documentclass{article}

\usepackage{microtype}
\usepackage{graphicx}
\usepackage{subcaption}
\usepackage{booktabs} 

\usepackage{hyperref}

\usepackage[preprint]{icml2026}

\usepackage{amsmath}
\usepackage{amssymb}
\usepackage{mathtools}
\usepackage{amsthm}

\usepackage{graphicx}
\usepackage{bbm}
\usepackage{seqsplit}
\usepackage{array}

\usepackage{enumitem}
\usepackage{hyperref}
\usepackage{url}

\usepackage[capitalize,noabbrev]{cleveref}

\theoremstyle{plain}

\theoremstyle{definition}

\theoremstyle{remark}

\usepackage[textsize=tiny]{todonotes}

\icmltitlerunning{Boosting In-Silicon Directed Evolution with Fine-Tuned Protein Language Model and Tree Search}

\begin{document}

\twocolumn[
  \icmltitle{Boosting In-Silicon Directed Evolution with Fine-Tuned Protein Language Model and Tree Search}



  \icmlsetsymbol{equal}{*}

  \begin{icmlauthorlist}
    \icmlauthor{Yaodong Yang}{cuhk}
    \icmlauthor{Yang Wang}{him}
    \icmlauthor{Jinpeng Li}{cuhk}
    \icmlauthor{Pei Guo}{him}
    \icmlauthor{Da Han}{him}
    \icmlauthor{Guangyong Chen}{him}
    \icmlauthor{Pheng-Ann Heng}{cuhk}
  \end{icmlauthorlist}

  \icmlaffiliation{cuhk}{Department of Computer Science and Engineering, The Chinese University of Hong Kong}
  \icmlaffiliation{him}{Hangzhou Institute of Medicine, Chinese Academy of Sciences}

  \icmlcorrespondingauthor{Jinpeng Li}{ljpadam@gmail.com}
  \icmlcorrespondingauthor{Guangyong Chen}{chenguangyong@him.cas.cn}

  \icmlkeywords{Machine Learning, ICML}

  \vskip 0.3in
]



\printAffiliationsAndNotice{}  

\begin{abstract}
Protein evolution through amino acid mutations is a cornerstone of life sciences. Recent advances in protein language models have shown rich evolutionary patterns, offering unprecedented potential for in-silicon directed evolution. However, existing directed evolution methods largely rely on heuristic evolution strategies and have yet to efficiently integrate the transformative protein language models with advanced optimization techniques, such as reinforcement learning, to adaptively learn superior evolution policies. To bridge this gap, we propose AlphaDE, a novel framework that evolves protein sequences by harnessing the innovative paradigms of large language models, such as fine-tuning and test-time inference. First, AlphaDE fine-tunes pretrained protein language models using masked language modeling on homologous protein sequences to activate the evolutionary plausibility of the interested protein family. Second, AlphaDE introduces test-time inference based on Monte Carlo tree search, which effectively evolves proteins with evolutionary guidance from the fine-tuned protein language model. Extensive benchmark experiments show that AlphaDE remarkably outperforms previous state-of-the-art methods even with few-shot fine-tuning. A case study further demonstrates that AlphaDE supports condensing the protein sequence space of avGFP through computational evolution.
\end{abstract}

\section{Introduction}\label{sec1}
Proteins are essential components of living systems, exhibiting a great diversity of functions among biological macromolecules \citep{holm_mapping_1996}. They play critical roles in a wide range of biochemical processes, including enzyme catalysis, cellular metabolism, immune responses, and signal transduction \citep{jiang_general_2024}. Protein engineering through directed evolution enables optimization of protein functions by seeking potential protein variants with improved properties such as the expression level and catalytic activity \citep{yang_machine-learning-guided_2019}. Traditional in vitro or in vivo experimental directed evolution approaches, such as deep mutational scanning \citep{fowler_deep_2014} and orthogonal DNA replication system \citep{ravikumar_orthogonal_2014}, directly measure the functional effects of protein mutations but are limited to exploring only a fraction of the possible protein space, and are usually expensive and laborious. To reduce the burden of the expensive wet experiments, recent advances in machine learning-guided approaches focus on building a surrogate sequence-function landscape \citep{yang_machine-learning-guided_2019}, and a lot of in-silicon directed evolution algorithms \citep{brookes_design_2018,sinai_adalead_2020,ren_proximal_2022,wang_self-play_2023} have been proposed to strategically explore the protein fitness landscape to identify desired sequence mutations via an iterative search process.

On the other hand, protein language models \citep{meier_language_2021,lin_evolutionary-scale_2023}, which encapsulate millions of years of evolutionary information through unsupervised pretraining in massive protein databases, are rapidly changing the domain of protein design \citep{madani_large_2023,jiang_rapid_2024,hayes_simulating_2025}. As protein language models implicitly learn complex evolutionary and structural dependencies from natural protein sequences, researchers have been increasingly employing protein language models for protein engineering tasks such as zero-shot inference of the functional effects of sequence substitutions to find high-fitness variations \citep{hie_efficient_2024,shanker_unsupervised_2024}. More recently, some pioneering works have begun to introduce pretrained protein language models in the process of directed evolution \citep{jiang_rapid_2024,yang_active_2025,tran_protein_2025}. However, although employing protein language models to recommend mutations, these works use simple heuristic search methods such as greedy selection \citep{jiang_rapid_2024} or beam search \citep{tran_protein_2025} for directed evolution. This leaves an important open question of how to seamlessly integrate protein language models into advanced optimization techniques, such as reinforcement learning, to adaptively learn superior evolution policies.

To bridge this gap, in this work, we propose a novel framework named AlphaDE to directly evolve protein sequences following the technical paradigm of natural large language models \citep{guo_deepseek-r1_2025}. Specifically, AlphaDE consists of a fine-tuning step and a Monte Carlo tree search (MCTS) inference step built on the protein language model. In the fine-tuning step, AlphaDE fine-tunes the pretrained protein language model with homologous protein sequences to activate its evolutionary plausibility for the interested protein family. In the MCTS-assisted inference step, similar to test-time MCTS in large language models to boost reasoning \citep{zhou_language_2024,guan_rstar-math_2025}, AlphaDE conducts an iterative tree search to efficiently optimize protein function via residue mutations guided by the fine-tuned protein language model. Through the two synergic steps, AlphaDE harvests superior directed evolution ability for proteins. To evaluate AlphaDE, we conduct computational benchmark experiments on eight distinct tasks. Impressively, AlphaDE substantially outperforms various in-silicon directed evolution methods. Further few-shot fine-tuning experiments reveal that AlphaDE's evolution ability can be activated by fine-tuning with only dozens of homologous sequences. Finally, in a proof-of-concept study on avGFP, we demonstrate that AlphaDE computationally condenses the functional sequence space, highlighting its broader applicability.

\section{Background}

\subsection{Protein Language Models}

In recent years, in analogy to large language models in natural language processing, protein language models such as ESM \citep{rives_biological_2021,lin_evolutionary-scale_2023}, ProteinBERT \citep{brandes_proteinbert_2022}, and ProGen \citep{madani_large_2023} have emerged as a transformative paradigm for modeling protein sequences. Pretrained by masked language modeling \citep{devlin_bert_2019} or autoregressive language modeling \citep{brown_language_2020} on evolutionary-scale protein databases, protein language models have demonstrated outstanding potential in protein design. For instance, recently, \citet{hayes_simulating_2025} prompt a protein language model (i.e., ESM3) to generate a bright fluorescent protein far different from known fluorescent proteins, which is estimated to simulate five hundred million years of evolution. Another notable example is that \citet{bhat_novo_2025} utilize contrastive language modeling to design peptide binders to conformationally diverse targets using only the amino acid sequence of the target protein.

Moreover, some works \citep{widatalla_aligning_2024,yang_steering_2025} utilize labeled sequence-fitness pairs to steer protein language models for \textit{de novo} protein sequence design to generate novel and high-quality sequences, showing the superior alignment ability of protein language models.

\subsection{Protein Directed Evolution}

Directed evolution is a classical paradigm for protein sequence design, where plenty of algorithms are developed to accelerate the in-silicon directed evolution process. AdaLead \citep{sinai_adalead_2020} is an advanced implementation of model-guided directed evolution with iteratively recombined and mutated operations for seed sequences. CMA-ES \citep{hansen_completely_2001} is a second-order evolutionary search algorithm that estimates the covariance matrix to adaptively adjust the search strategy of the upcoming generations. Bayesian optimization (BO) \citep{snoek_practical_2012} is a classical paradigm for the sequential design problem \citep{mockus_bayesian_1989}, which estimates the uncertainty and constructs the acquisition function for exploration. DbAS \citep{brookes_design_2018} establishes a probabilistic framework that trains a variational autoencoder (VAE) \citep{kingma_auto-encoding_2022} to model the distribution of high-fitness sequences and adaptively samples sequences from this trained VAE to explore the fitness landscape. Following DbAS, CbAS \citep{brookes_conditioning_2019} estimates the probability distribution conditioned on the desired properties with model-based adaptive sampling and additionally considers a regularization to stabilize the model-guided search process. DyNA-PPO \citep{angermueller_model-based_2020} formulates protein sequence design as a sequential decision-making problem and uses proximal policy optimization (PPO) \citep{schulman_proximal_2017} to perform sequence generation. PEX \citep{ren_proximal_2022} aims to search for effective candidates of low-order mutants near the wild-type, and formulates this process as a proximal optimization problem to solve. EvoPlay \citep{wang_self-play_2023} uses the self-play reinforcement learning inspired by AlphaZero \citep{silver_general_2018} to optimize protein sequences. Recently, TreeNeuralTS and TreeNeuralUCB \citep{qiu_tree_2024} combine tree search with bandit machine learning for directed evolution, which expands a tree starting from the initial sequence with the guidance of a bandit machine learning model. Specifically, TreeNeuralTS adopts Thompson sampling \citep{thompson_likelihood_1933} while TreeNeuralUCB utilizes the upper confidence bound \citep{auer_finite-time_2002} to explore the sequence space.

More recently, with the emergence and development of protein language models, some pioneering works explore integrating pretrained protein language models in the process of directed evolution \citep{jiang_rapid_2024,yang_active_2025,tran_protein_2025}. For example, \citet{jiang_rapid_2024} and \citet{yang_active_2025} present the multi-round active learning to rapidly improve protein activity, which alternates between collecting sequence fitness using a wet-lab assay and training a protein language model to prioritize new sequences to screen in the next round. Meanwhile, LatentDE \citep{tran_latentde_2025} uses a gradient-based method to search in the latent space represented by VAE with protein embeddings extracted from a frozen, general-purpose pretrained protein language model (i.e., ESM-2). Moreover, MLDE \citep{tran_protein_2025} introduces an optimization pipeline that utilizes protein language models to pinpoint the mutation hotspots and then suggest replacements by heuristically calculating a k-mer's relevancy and entropy in a sequence. However, despite leveraging protein language models to recommend mutations, the above works use simple search methods such as greedy selection \citep{jiang_rapid_2024} or beam search \citep{tran_protein_2025}, thus leaving a critical challenge unexplored in protein directed evolution: how to integrate protein language models with advanced optimization techniques like reinforcement learning to efficiently boost protein evolution. To bridge this gap, here we propose AlphaDE with a principle of effective searching (i.e., AlphaZero-like MCTS) with strong prior guidance (i.e., fine-tuned protein language model), inspired by the natural large language model technique paradigm where fine-tuning \citep{shao_deepseekmath_2024} and test-time inference \citep{guan_rstar-math_2025} improve problem-solving, to establish a new paradigm for in-silicon protein directed evolution.

\section{AlphaDE}

\begin{figure*}[htbp]
    \centering
    \includegraphics[width=1.0\linewidth]{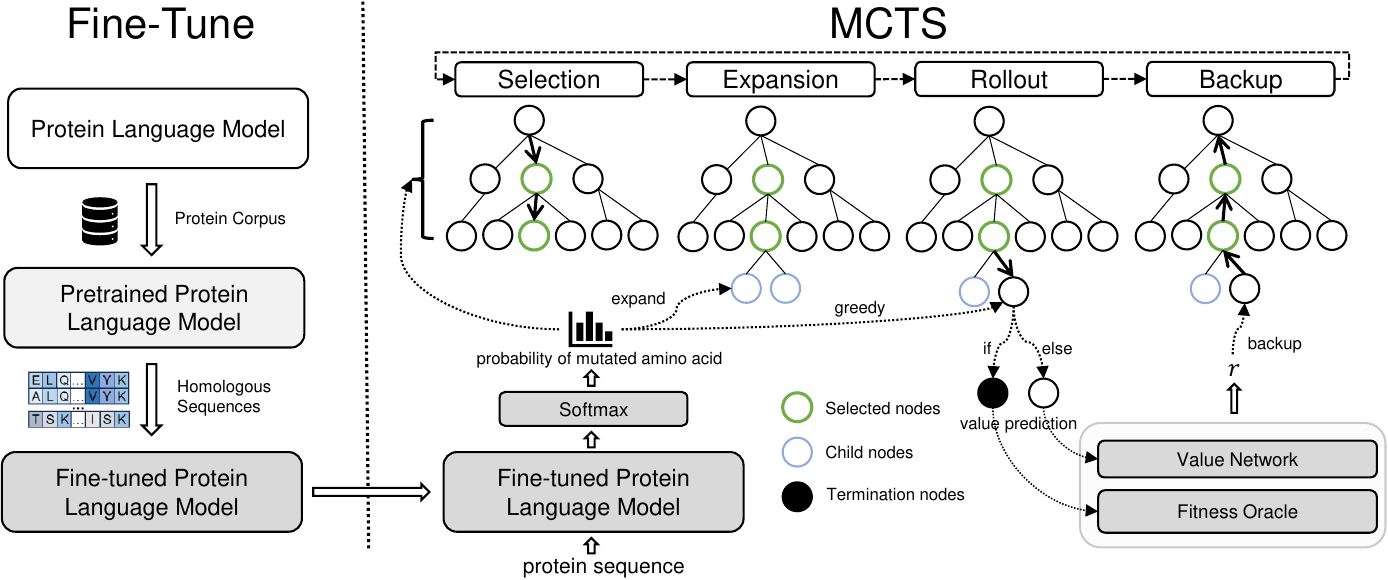}
    \caption{The framework of AlphaDE. It consists of a fine-tuning step and an MCTS inference step.}
    \label{fig:framework}
\end{figure*}

In this section, we present AlphaDE for protein directed evolution in detail. First, we give the problem definition of the directed evolution in Section~\ref{section:problem_definition}. Second, in Section~\ref{section:finetuing_step}, we introduce how to fine-tune the pretrained protein language model to give prior guidance on the next mutation residues. Third, in Section~\ref{section:mcts_step}, we show how to perform tree search to directly evolve protein sequences following the mutation guidance from the fine-tuned protein language model. The whole framework of AlphaDE is shown in Figure~\ref{fig:framework}.

\subsection{Problem Definition}
\label{section:problem_definition}
Protein directed evolution can be formulated as a Markov decision process (MDP) \citep{bellman_markovian_1957}, given that the next mutation residue to be chosen depends on the current protein sequence \citep{wang_self-play_2023}. The MDP is defined as $M=(S, A, P, R)$ where $S$ denotes the set of states that describe the current protein sequence, $A$ denotes the set of actions that indicate the chosen position and residue type to be mutated from the current protein sequence, and $P:S \times A \rightarrow S$ is the state transition function where the current protein sequence incorporates the chosen residue position and type to mutate to a new protein sequence. The protein sequence state $s \in S$ is represented by a binary matrix of size $20 \times L$ with columns indicating positions ($L$) and rows indicating amino acids (20). Correspondingly, the action $a \in A$ is a flattened one-hot vector of size $20 \times L$. When an action $a$ is performed, the element of value 1 in the action vector is selected, resulting in a single-site mutation on the current protein sequence $s$. The element indicated by the mutation action in the state matrix is changed to 1, while the other elements (residue types) of the same column (position in protein) are all set to 0. An episode contains a series of mutation actions $a_{t}$ and states $s_{t}$ where $0 < t < T$ and $T$ is the termination step. $R: S \rightarrow \mathbb{R}$ is the episodic reward function indicating protein fitness, such as binding affinity, which can be accessed by an experimental assay or predicted by a surrogate oracle. Directed evolution aims to take actions that maximize $\overline{r}$, which is approximated under repeated rollouts \citep{gelly_monte-carlo_2011} as
\begin{equation}
\label{eq:q}
    \overline{r}(s,a) = \frac{1}{N(s,a)} \sum_{j=1}^{N(s)} \mathbbm{I}_{j}(s,a)r_{j}(s),
\end{equation}
where $N(s)$ denotes the number of rollouts starting from state $s$ and $N(s, a)$ is the visit count that action $a$ has been taken from state $s$. $\mathbbm{I}_{j}(s, a)$ is an indicator function with value 1 if action $a$ is selected from state $s$ at the $j$th rollout round, 0 otherwise. $r_{j}(s)$ is the episodic reward of the final protein sequence at the terminal state for the $j$th rollout round starting from state $s$. A larger $\overline{r}(s, a)$ indicates a higher expected episodic reward by taking action $a$ from state $s$. The measurement of episodic reward $r_{j}(s)$ traditionally relies on wet experiments, which are time-consuming and expensive. Therefore, in-silicon directed evolution simulates the ground-truth protein fitness landscape by a proxy oracle model to replace the wet-lab measurements.

\subsection{Fine-Tuning Pretrained Protein Language Model}
\label{section:finetuing_step}
Fine-tuning natural large language models on domain-specific data has become a standard practice to better solve specialized tasks, such as math \citep{shao_deepseekmath_2024}, coding \citep{muennighoff2024octopack}, and medical analysis \citep{christophe2024med}, by adapting large language models to enhance task-specific capabilities. Inspired by this, we fine-tune the protein language model with homologous protein sequences to activate its evolutionary plausibility for the interested protein family. Specifically, AlphaDE uses the unsupervised masked language modeling objective \citep{devlin_bert_2019} without fitness labels. During fine-tuning, each input protein sequence is corrupted by replacing a fraction of amino acids with a special mask token. The protein language model is then trained to predict masked tokens in the corrupted sequences sampled from a homologous sequence set $S_{h}$:
\begin{equation}
    L_{mlm}=\mathbb{E}_{s \sim S_{h}} \mathbb{E}_{M} \sum_{i \in M} -\\log p(s_{i} | s_{/M}).
\end{equation}
For each sequence $s$, we sample a set of indices $M$ to mask, replacing the true token at each index $i$ with the mask token. For each masked token, we independently minimize the negative log likelihood of the true amino acid $s_{i}$, given the masked sequence $s_{/M}$ as context. Intuitively, to predict a masked residue, the protein language model must identify dependencies between the masked site and the unmasked parts of the sequence, therefore capturing the co-evolution information among residues. After the fine-tuning step, the protein language model recovers masked protein sequences towards the possible existing sequences of the same family as homologous sequences, which gives a prior distribution of next mutations for the next MCTS step.

\subsection{MCTS with Fine-Tuned Protein Language Model}
\label{section:mcts_step}
We use Monte Carlo tree search (MCTS) to solve the MDP of directed evolution with guidance from protein language models, inspired by the test-time reasoning improvements in large language models that MCTS-based search generates better solutions \citep{zhou_language_2024,guan_rstar-math_2025}. Diverging from prior MCTS works that train neural networks on expert data from scratch for action guidance \citep{rosin_multi-armed_2011,silver_mastering_2016}, our approach leverages a fine-tuned protein language model to provide evolutionary guidance for mutation decisions, which predicts the next mutation action (i.e., the residue position and residue type) given the current protein sequence. Specifically, at each step, the fine-tuned protein language model receives the protein sequence and outputs the probability logits of each residue type at each residue position to guide the MCTS search.

When mutating protein sequences with the fine-tuned protein language model, although we could mutate a protein sequence in a greedy manner by taking the next mutation with the maximum probability, it is prone to being stuck in a local optimum due to the unpredictable complexity of the protein fitness landscape \citep{romero_exploring_2009}. Moreover, the predicted mutation with the maximum probability does not mean that it is optimal because the protein language model is not optimized for a specific protein fitness. To directly maximize a given protein fitness, we integrate MCTS into AlphaDE to solve the MDP of directed evolution with the help of a fine-tuned protein language model for mutation guidance. Next, we explain the MCTS part of AlphaDE in the background of protein directed evolution.

MCTS \citep{browne_survey_2012} adopts a tree structure to perform simulation iterations and estimate the state value of actions. Meanwhile, it uses the previously estimated action values to guide the search process towards higher rewards. As shown in Figure~\ref{fig:framework}, the MCTS part in AlphaDE consists of the following four steps per iteration:

\begin{itemize}[leftmargin=*]

\item \textbf{Selection}. 
Each iteration starts from sequence $s_{\tau}$, recursively selecting the best child node until reaching a leaf node $a_{\tau + l}$, i.e., a node that has not been expanded or terminated, after $l$ selections.
At each selection step $t \in [1, l]$, a selection criterion determines the best child node to be chosen, which balances exploitation and exploration to avoid being trapped in local optimum. We use Predictor with Upper Confidence bounds applied to Trees (PUCT) \citep{rosin_multi-armed_2011} as the selection criterion for each candidate child node as
\begin{equation}
\label{eq:puct}
    U_{puct}(s_{\tau+t-1}, a) = \frac{W_{a}}{N_{a}} + c P_{plm}(a \vert s_{\tau+t-1}) \frac{\sqrt{N}}{1+N_{a}},
\end{equation}
where a constant $c$ controls the exploration degree. $W_{a}$ is the cumulative reward of node $a$, $N$ is the total visit count, and $N_{a}$ is the visit count for node $a$. The term $\frac{\sqrt{N}}{1+N_{a}}$ encourages selection of less-visited nodes, while $P_{plm}(a|s_{\tau+t-1})$ favors beneficial mutations indicated by the fine-tuned protein language model. On the other hand, $\frac{W_{a}}{N_{a}}$ encourages exploitation of the nodes with high average protein fitness.

\begin{table*}[htbp]
\small
\centering
\caption{Benchmarking in-silicon directed evolution methods. We present the maximum fitness obtained in 1000 black-box oracle queries. Results are presented as mean with standard deviation over five independent trials.}
\begin{tabular}{lllllllll}
\hline
Method & avGFP & AAV & TEM & E4B & AMIE & LGK & PAB1 & UBE2I \\ \hline
AlphaDE & \textbf{3.86}{\tiny$\pm$0.00} & \textbf{7.95}{\tiny$\pm$1.09} & \textbf{1.22}{\tiny$\pm$0.01} & \textbf{7.75}{\tiny$\pm$0.26} & \textbf{0.24}{\tiny$\pm$0.01} & \textbf{0.04}{\tiny$\pm$0.00} & \textbf{1.47}{\tiny$\pm$0.59} & 2.97{\tiny$\pm$0.03} \\ \hline
LatentDE {\tiny\citep{tran_latentde_2025}} & 3.79{\tiny$\pm$0.00} & 4.72{\tiny$\pm$1.26} & 1.20{\tiny$\pm$0.00} & 3.88{\tiny$\pm$1.89} & -3.34{\tiny$\pm$0.05} & -1.13{\tiny$\pm$0.26} & 0.58{\tiny$\pm$0.35} & 1.29{\tiny$\pm$0.00} \\ \hline
MLDE {\tiny\citep{tran_protein_2025}} & 2.04{\tiny$\pm$0.62} & -3.38{\tiny$\pm$0.18} & 0.08{\tiny$\pm$0.02} & 3.84{\tiny$\pm$1.53} & -1.38{\tiny$\pm$1.82} & -0.19{\tiny$\pm$0.39} & 0.92{\tiny$\pm$0.18} & 2.75{\tiny$\pm$0.15} \\ \hline
TreeNeuralTS {\tiny\citep{qiu_tree_2024}} & 2.44{\tiny$\pm$0.78} & 2.47{\tiny$\pm$0.98} & 0.27{\tiny$\pm$0.06} & 0.79{\tiny$\pm$0.21} & -0.22{\tiny$\pm$0.60} & \textbf{0.04}{\tiny$\pm$0.00} & 1.02{\tiny$\pm$0.46} & \textbf{2.98}{\tiny$\pm$0.01} \\ \hline
TreeNeuralUCB {\tiny\citep{qiu_tree_2024}} & 2.37{\tiny$\pm$0.75} & 3.85{\tiny$\pm$1.55} & 0.19{\tiny$\pm$0.08} & 0.70{\tiny$\pm$0.25} & -0.19{\tiny$\pm$0.49} & \textbf{0.04}{\tiny$\pm$0.00} & 1.27{\tiny$\pm$0.47} & 2.95{\tiny$\pm$0.04} \\ \hline
PEX (MuFacNet) {\tiny\citep{ren_proximal_2022}} & 3.01{\tiny$\pm $0.99} & 4.07{\tiny$\pm $1.52} & 0.22{\tiny$\pm $0.12} & 0.91{\tiny$\pm $0.06} & 0.20{\tiny$\pm $0.03} & \textbf{0.04}{\tiny$\pm $0.00} & 1.38{\tiny$\pm $0.25} & 2.96{\tiny$\pm $0.04} \\ \hline
PEX (CNN) {\tiny\citep{ren_proximal_2022}} & 3.30{\tiny$\pm $0.78} & 3.30{\tiny$\pm $1.97} & 0.18{\tiny$\pm $0.04} & 0.69{\tiny$\pm $0.16} & 0.07{\tiny$\pm $0.17} & 0.03{\tiny$\pm $0.00} & 1.18{\tiny$\pm $0.15} & 2.95{\tiny$\pm $0.07} \\ \hline
EvoPlay {\tiny\citep{wang_self-play_2023}} & 1.72{\tiny$\pm $0.16} & -3.45{\tiny$\pm $1.47} & 0.01{\tiny$\pm $0.01} & -0.40{\tiny$\pm $0.37} & -0.88{\tiny$\pm $0.91} & -1.09{\tiny$\pm $0.19} & 0.34{\tiny$\pm $0.17} & 1.87{\tiny$\pm $0.89}  \\ \hline
AdaLead {\tiny\citep{sinai_adalead_2020}} & 1.81{\tiny$\pm $0.11} & -2.37{\tiny$\pm $0.30} & 0.10{\tiny$\pm $0.06} & 0.31{\tiny$\pm $0.26} & -7.83{\tiny$\pm $0.91} & -1.36{\tiny$\pm $0.10} & 0.63{\tiny$\pm $0.07} & 0.29{\tiny$\pm $0.60} \\ \hline
DyNA-PPO {\tiny\citep{angermueller_model-based_2020}} & 1.57{\tiny$\pm $0.02} & -3.53{\tiny$\pm $0.12} & 0.02{\tiny$\pm $0.00} & -0.20{\tiny$\pm $0.15} & -7.96{\tiny$\pm $0.22} & -1.29{\tiny$\pm $0.03} & 0.45{\tiny$\pm $0.07} & -0.04{\tiny$\pm $0.02} \\ \hline
DbAS {\tiny\citep{brookes_design_2018}} & 2.14{\tiny$\pm $0.83} & -3.45{\tiny$\pm $0.26} & 0.02{\tiny$\pm $0.01} & -0.44{\tiny$\pm $0.10} & -2.33{\tiny$\pm $2.80} & -0.43{\tiny$\pm $0.61} & 0.37{\tiny$\pm $0.08} & 1.92{\tiny$\pm $0.55} \\ \hline
CbAS {\tiny\citep{brookes_conditioning_2019}} & 1.60{\tiny$\pm $0.09} & -3.46{\tiny$\pm $0.28} & 0.02{\tiny$\pm $0.00} & -0.27{\tiny$\pm $0.25} & -1.94{\tiny$\pm $1.51} & -1.35{\tiny$\pm $0.02} & 0.43{\tiny$\pm $0.13} & 2.25{\tiny$\pm $0.80} \\ \hline
BO {\tiny\citep{snoek_practical_2012}} & 1.57{\tiny$\pm $0.03} & -3.84{\tiny$\pm $0.16} & 0.02{\tiny$\pm $0.00} & -0.33{\tiny$\pm $0.26} & -5.71{\tiny$\pm $2.25} & -1.45{\tiny$\pm $0.05} & 0.39{\tiny$\pm $0.09} & 0.02{\tiny$\pm $0.13} \\ \hline
CMA-ES {\tiny\citep{hansen_completely_2001}} & 1.60{\tiny$\pm $0.03} & -3.50{\tiny$\pm $0.19} & 0.02{\tiny$\pm $0.00} & -0.09{\tiny$\pm $0.19} & -7.92{\tiny$\pm $0.20} & -1.29{\tiny$\pm $0.00} & 0.53{\tiny$\pm $0.05} & -0.01{\tiny$\pm $0.04} \\ \hline
\end{tabular}
\label{tab:benchmark}
\end{table*}

\item \textbf{Expansion}.
Given a selected leaf node $a_{\tau + l}$, the fine-tuned protein language model computes the probability $P_{plm}(a \vert s_{\tau+l})$ for each expandable action $a \in A$ as a prior distribution over mutations. Here, $s_{\tau+l}$ is the state context of node $a_{\tau + l}$, and $A$ denotes the legal action space, i.e., the residue type and position in protein. The expanded child nodes of the leaf node $a_{\tau + l}$ are immediately added and initialized in the tree.

\item \textbf{Rollout}. 
The value of the reached leaf node $a_{\tau+l}$ is evaluated by a rollout. From the leaf node, MCTS recursively mutates the sequence until termination at step $T$. Then the final sequence $s_T$ is evaluated by a fitness oracle for reward $r$. During the rollout, each mutation is selected greedily. To speed up the computationally expensive rollout process, AlphaDE trains a value network to predict the state value of the reached leaf nodes as AlphaZero \citep{silver_general_2018}. Specifically, if the reached leaf node $s_{\tau+l}$ is a termination node, it is evaluated by the fitness oracle. Otherwise, it is predicted by the value network.

\item \textbf{Backup}. After rollout, the final reward $r$ is backpropagated along the visited nodes to update their statistics until the root node. The detailed updating for tree nodes is elaborated as
\begin{equation}
\label{eq:update}
    N_{a} \leftarrow N_{a} + 1, W_{a} \leftarrow W_{a} + r, a \leftarrow \mbox{parent of } a,
\end{equation}
where $N_{a}$ is the visit count and $W_{a}$ is the cumulative reward of node $a$. For each selection $t \in [1, l]$, the statistics of node $a_{\tau+t}$ are updated by adding the rollout reward of $a_{\tau+l}$ to $W_{a}$ and by increasing $N_{a}$ by 1.
\end{itemize}

\section{Experiments}

In this section, we conduct comprehensive experiments to validate the effectiveness of AlphaDE. First, in Section~\ref{section:benchmark}, we benchmark AlphaDE with various directed evolution algorithms on extensive tasks. Second, we study AlphaDE under different settings of homology sequence availability, such as few-shot fine-tuning in Section~\ref{section:fewshot_finetuning} and zero-shot scaling with model sizes in Section~\ref{section:zeroshot_scaling}. Finally, we utilize AlphaDE to computationally condense the protein sequence space of the fluorescent protein in Section~\ref{section:avgfp_condesing}. The ablation study of AlphaDE is provided in Appendix~\ref{appendix:ablation}.

\subsection{Benchmark experiments}
\label{section:benchmark}

\begin{table*}[htbp]
\centering
\caption{Results of AlphaDE with ProtBert and ESM-1b models. We present the maximum fitness obtained in 1000 black-box oracle queries. Results are averaged over five independent trials.}
\begin{tabular}{lllllllll}
\hline
AlphaDE Model & avGFP & AAV & TEM & E4B & AMIE & LGK & PAB1 & UBE2I \\ \hline
fine-tuned ProtBert \citep{elnaggar_prottrans_2022} & 3.09 & 7.66 & 0.49 & 7.66 & 0.00 & -0.01 & 0.41 & 2.71 \\
w/o fine-tuning & 1.53 & -0.94 & 0.24 & -0.37 & 0.02 & 0.00 & 0.19 & 1.36 \\ \hline
fine-tuned ESM-1b \citep{rives_biological_2021} & 3.09 & 16.97 & 0.49 & 7.85 & 0.03 & 0.01 & 0.60 & 1.49 \\
w/o fine-tuning & 1.46 & -0.96 & 0.24 & 2.10 & -0.53 & 0.00 & 0.42 & 1.47 \\ \hline
\end{tabular}
\label{tab:otherplms}
\end{table*}

\begin{figure*}[ht]
    \centering
    \includegraphics[width=1.0\linewidth]{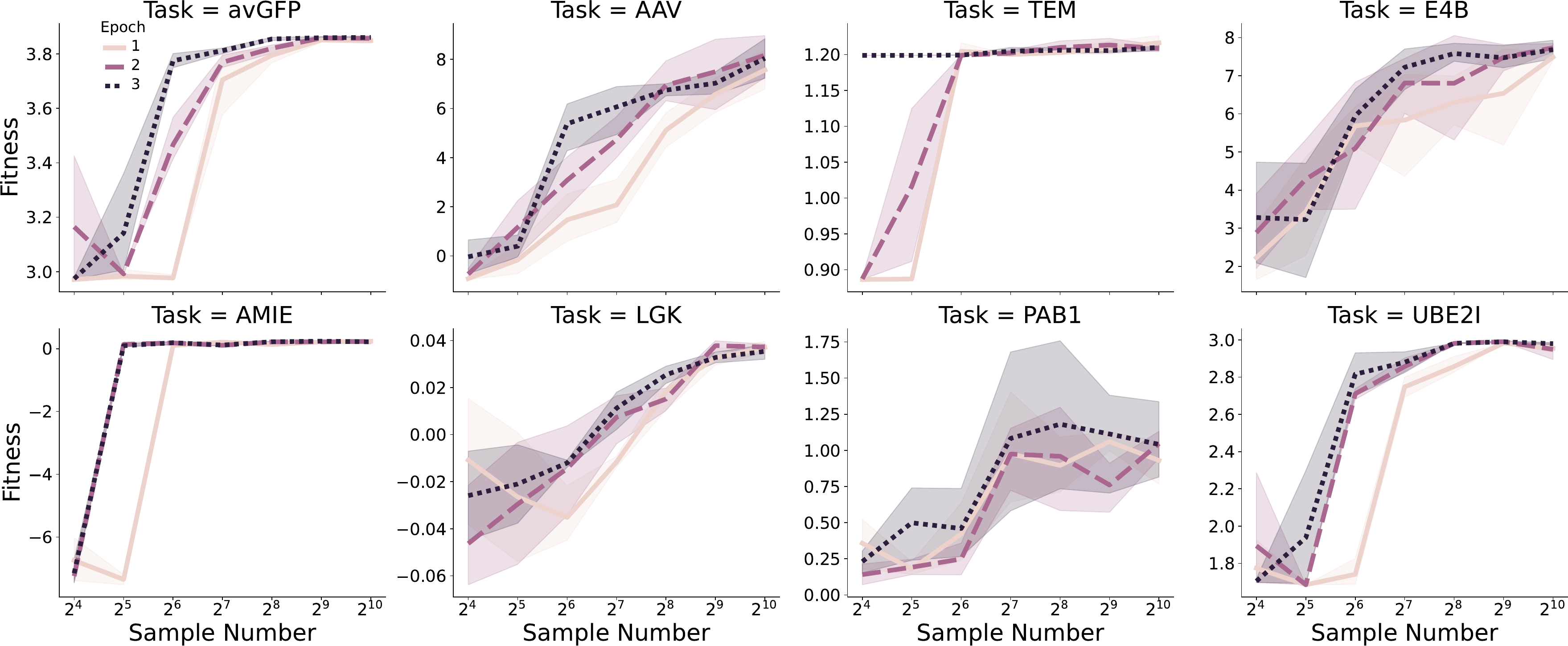}
    \caption{AlphaDE with fine-tuned ESM2-35M, which are fine-tuned with different numbers of sequences randomly sampled from the whole data distribution. 95\% confidence intervals are shadowed.}
    \label{fig:fewshot_percent100}
\end{figure*}

\begin{figure*}[ht]
    \centering
    \includegraphics[width=1.0\linewidth]{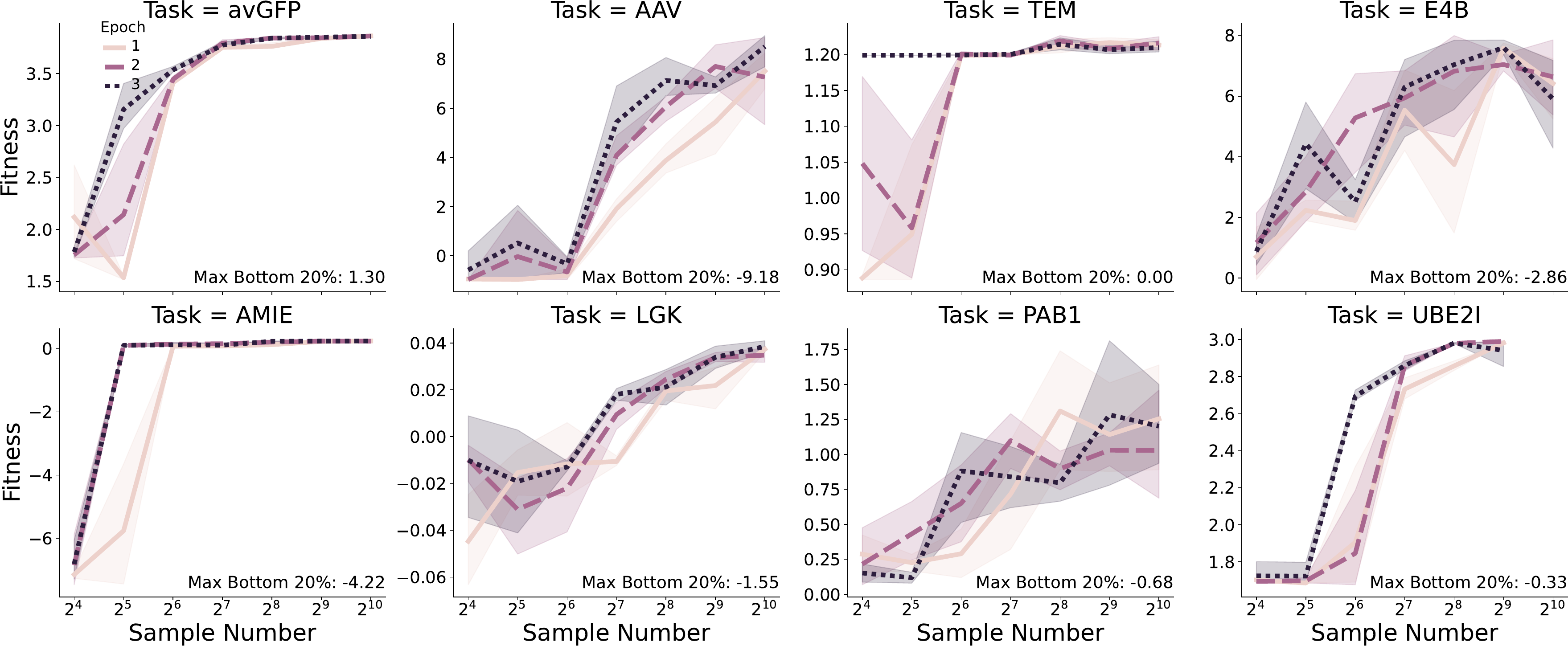}
    \caption{AlphaDE with fine-tuned ESM2-35M models of different numbers of sequences randomly sampled from the bottom 20\% data. The ``Max Bottom 20\%'' denotes the maximum fitness value in the bottom 20\% data. 95\% confidence intervals are shadowed.}
    \label{fig:fewshot_percent20}
\end{figure*}

\begin{figure*}[htb]
    \centering
    \includegraphics[width=1.0\linewidth]{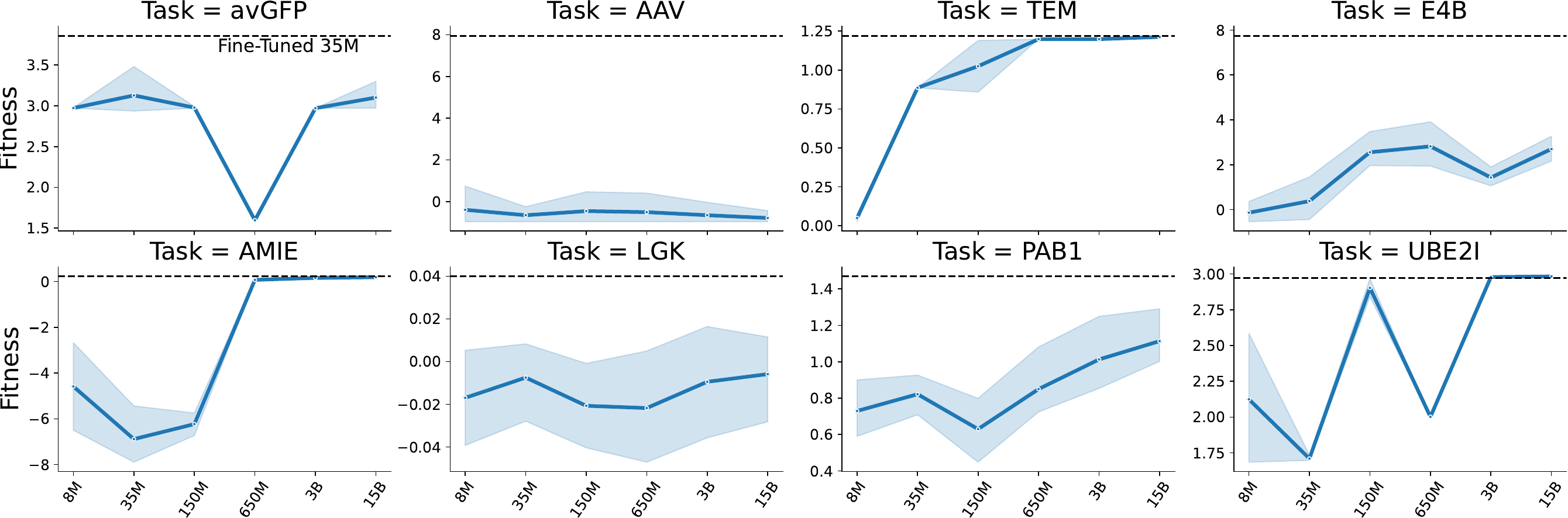}
    \caption{AlphaDE's performance scales with the sizes of pretrained protein language models. The black horizontal dashed line indicates AlphaDE with fine-tuned ESM2-35M. 95\% confidence intervals are shadowed.}
    \label{fig:pretrained_scaling_law}
\end{figure*}

\begin{figure*}[htb]
    \centering
    \includegraphics[width=1.0\linewidth]{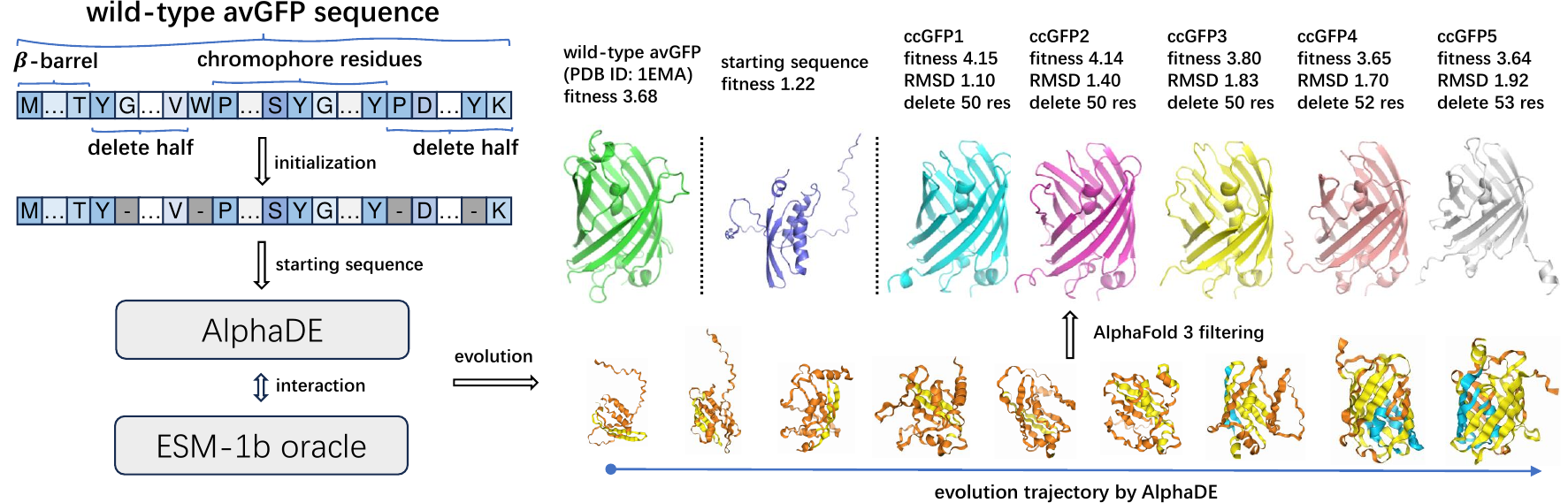}
    \caption{The illustrated process of AlphaDE to condense the sequence space of avGFP. The evolution trajectory is sampled during one trial, and the first structure is the prediction of the starting sequence by AlphaFold 3.}
    \label{fig:avgfp_condensing}
\end{figure*}

We benchmark AlphaDE and baselines on a suite of eight in-silicon protein engineering tasks \citep{ren_proximal_2022}. These tasks involve extensive protein applications such as biosensor design and industrial enzyme renovation. More descriptions about the eight benchmark tasks are provided in Appendix~\ref{appendix:benchmark_tasks}. Following previous works \citep{ren_proximal_2022,wang_self-play_2023,qiu_tree_2024}, we simulate the ground-truth fitness landscape of protein with an oracle model TAPE \citep{rao_evaluating_2019} to replace wet laboratory measurements.

We compare AlphaDE with various baselines including LatentDE, MLDE, TreeNeuralTS, TreeNeuralUCB, PEX, EvoPlay, AdaLead, DyNA-PPO, DbAS, CbAS, BO, and CMA-ES. We follow the officially released codebase and configurations of LatentDE, MLDE, TreeNeuralTS, TreeNeuralUCB, PEX (one version with a CNN model and another with a MuFacNet model), and EvoPlay for running. We run AdaLead, DyNA-PPO, DbAS, CbAS, BO, and CMA-ES with the implementations in FLEXS \citep{sinai_adalead_2020}, following the same benchmark setting, such as the starting sequence and fitness oracle as Ren et al. \citep{ren_proximal_2022}. We average the results of AlphaDE and all baselines over 5 independent trials. For all methods, we set the same oracle budget of 1000 queries. For AlphaDE, here we use the pretrained ESM2-series models \citep{lin_evolutionary-scale_2023} for fine-tuning due to its strong performance, wide adoption, and availability in various model sizes, which is convenient for our study. Specifically, in the standard AlphaDE, we use ESM2-35M for fine-tuning due to its efficiency in model size and effectiveness in evolution plausibility. Details of the homologous sequence datasets for fine-tuning are in Appendix~\ref{appendix:dataset}, while hyperparameters of AlphaDE are given in Appendix~\ref{appendix:hyperparameter}. For each task, the starting protein sequence to evolve is the sequence with the lowest protein fitness. The benchmark results for eight tasks are summarized in Table~\ref{tab:benchmark}, which includes mean values and standard deviations.

The results in Table~\ref{tab:benchmark} show that AlphaDE outperforms various baselines in most tasks. For example, in the task of AAV, the second-best method, MLDE, obtains a fitness value of 4.72 while AlphaDE achieves 7.95 with an improvement of 68.43\%. A similar situation also happens in the task of E4B, which exhibits AlphaDE's great evolution ability by combining effective searching (i.e., AlphaZero-like MCTS) with strong prior guidance (i.e., fine-tuned protein language model). At the same time, we also examine the diversity of evolved sequences by AlphaDE in Appendix~\ref{appendix:seq_diversity}, which shows nearly 100\% sequence diversity across running trials. AlphaDE with different ESM2 model sizes is tested in Appendix~\ref{appendix:esm2size}. Meanwhile, the ablation study of AlphaDE on the fine-tuning step and MCTS inference step is given in Appendix~\ref{appendix:ablation}. Benchmark results using another oracle based on ESM-1b further validate AlphaDE as shown in Appendix~\ref{appendix:esm1b_oracle}. Moreover, the study of hyperparameter sensitivity of AlphaDE is available in Appendix~\ref{appendix:hyperparameter_sensitivity}, while AlphaDE's computation efficacy is provided in Appendix~\ref{appendix:efficacy}.

We also test AlphaDE's compatibility with other protein language models such as ProtBert \citep{elnaggar_prottrans_2022} (model size 420M) and ESM-1b \citep{rives_biological_2021} (model size 650M), which are trained with the standard BERT \citep{devlin_bert_2019} architecture using the UniRef100 \citep{suzek_uniref_2007} database and with the optimized RoBERTa \citep{liu_roberta_2019} architecture using the UniRef50 \citep{suzek_uniref_2007} database, respectively. The results of AlphaDE with ProtBert and ESM-1b are shown in Table~\ref{tab:otherplms}. Generally, AlphaDE with the two different protein language models achieves superior performance, demonstrating AlphaDE's compatibility. On the other hand, AlphaDE with ProtBert and ESM-1b perform differently in some tasks, suggesting the evolution ability of protein language models varies in different protein families. Meanwhile, the fine-tuning step is the key to boosting performance, as directly using the pretrained versions of models leads to huge fitness loss.

\subsection{Few-Shot Fine-Tuning Experiments}
\label{section:fewshot_finetuning}
One of the biggest advantages of large language models is that they are few-shot learners \citep{brown_language_2020}. Here we investigate whether this advantage is held in the protein language model for directed evolution tasks. We use the ESM2-35M model as the base model for few-shot fine-tuning and fine-tune it with different numbers of protein sequences including 16, 32, 64, 128, 256, 512, and 1024. These protein sequences for few-shot fine-tuning are randomly sampled from the whole dataset and are used to fine-tune ESM2-35M with 3 epochs. The results of AlphaDE with different fine-tuned ESM2-35M models of different sequence numbers are given in Figure~\ref{fig:fewshot_percent100}. We see that, even with dozens of protein sequences, the evolution ability of AlphaDE could be greatly improved, and the performance increases with the number of fine-tuning protein sequences. 

Although AlphaDE does not utilize fitness labels of protein sequences in the fine-tuning and MCTS inference steps, we additionally test whether the protein fitness influences AlphaDE's performance by fine-tuning on the sampled sequences from the bottom 20\% of data with the lowest fitness. The results of fine-tuning with low-fitness data are plotted in Figure~\ref{fig:fewshot_percent20}. The curves are very similar to Figure~\ref{fig:fewshot_percent100}, showing that AlphaDE's evolution capability is not affected by fitness values during fine-tuning. This suggests applicability in real-world settings where only sequences are available, without costly wet measurement assays.
We also show that AlphaDE works with the sequences from homology searching such as HMMER \citep{potter_hmmer_2018} in Appendix~\ref{appendix:phmmer}, where homology sequences are retrieved from a database.

\subsection{Zero-Shot Scaling Experiments on Model Size}
\label{section:zeroshot_scaling}

Next, we study how AlphaDE scales with pretrained protein language model sizes in a zero-shot setting, testing ESM2 models from 8M to 15B. Results in Figure~\ref{fig:pretrained_scaling_law} show that, in most tasks except avGFP and AAV, AlphaDE's performance roughly increases with pretrained model sizes. This is different from Table~\ref{tab:esm2size} in Appendix~\ref{appendix:esm2size}, where the fine-tuned model size has little effect. Interestingly, with the largest pretrained 15B model, AlphaDE's performance matches its fine-tuned ESM2-35M version on tasks of TEM, AMIE, and UBE2I. However, in most tasks, the pretrained models, even 15B, perform worse than the standard AlphaDE with fine-tuned ESM2-35M. We see two points here. First, larger pretrained models encode more evolutionary information, enhancing AlphaDE's evolution ability in most cases. Second, the fine-tuned small models with homologous sequences provide strong task-specific evolution ability, underscoring the necessity of fine-tuning in AlphaDE's pipeline.

\subsection{Condensing Protein Sequence Space}
\label{section:avgfp_condesing}

The origin and evolution of protein folds remain a fundamental challenge in biology \citep{chothia_how_1997,levitt_nature_2009}, with methods for exploring evolutionary trajectories still underdeveloped. Leveraging AlphaDE's evolutionary capability, we address this through condensing the avGFP sequence space by evolving an incomplete, non-folding sequence into a functional, folded protein. Inspired by hypotheses that proteins evolve from small random peptides into complex folded structures \citep{nepomnyachiy_complex_2017,kolodny_bridging_2021}, we keep the chromophore and $\beta$-barrel residues of avGFP, mask half the remaining sequence, and use AlphaDE to optimize the predicted fluorescence fitness to recover protein function. If AlphaDE recovers the predicted fluorescence function with fewer amino acid residues than the wild-type avGFP sequence, it successfully condenses the protein sequence space through computational evolution. Fluorescence intensity is predicted using an ESM-1b-based oracle landscape simulator \citep{ren_proximal_2022}, where character `-' denotes residue deletion. The computational evolution process is illustrated in Figure~\ref{fig:avgfp_condensing}.

We conduct 10 condensing trials and select the top 2 sequences from each trial with the highest fitness while keeping the number of deleted residues no less than 50. Then we use AlphaFold 3 \citep{abramson_accurate_2024} to predict structures of these 20 sequences and align their structures with the wild-type avGFP (PDB ID: 1EMA). Finally, we filter the top 5 sequences (ccGFP1-5) with the smallest RMSD to the wild-type structure, as in Figure~\ref{fig:avgfp_condensing}. As we see, AlphaDE successfully evolves the truncated, non-folding sequence to predictedly functional, folded proteins with fewer residues than the wild-type avGFP. More details, including the comparison to AlphaDE without the fine-tuning step and the amino acid residues of ccGFP1-5's protein sequences, can be referred to in Appendix~\ref{appendix:avgfp}.

\section{Conclusion}
In this work, we propose AlphaDE to boost the in-silicon protein directed evolution. Following the paradigm of natural large language models, AlphaDE consists of a fine-tuning step and an MCTS test-time inference step based on the protein language models to attain superior evolution ability. Specifically, AlphaDE conducts MCTS to efficiently evolve proteins via residue mutations guided by the fine-tuned protein language model. Benchmark experiments on eight tasks show that AlphaDE achieves state-of-the-art performance compared with various baselines, even in the few-shot setting. Additionally, we demonstrate that AlphaDE supports computationally condensing the protein sequence space.

For future work, on the one hand, integrating large natural language models into AlphaDE to provide the explainability of the evolution process is interesting. On the other hand, applying AlphaDE to industrial applications, such as improving the activity of enzymes, has great economic prospects. Third, extending AlphaDE to the setting of multi-objective directed evolution is also a promising direction.




\section*{Impact Statement}




Directed evolution is a powerful computational tool for protein engineering. The proposed AlphaDE significantly boosts the efficiency of in-silicon directed evolution, which exhibits great potential for real-world protein engineering applications. At the same time, we emphasize safety concerns that it can be misused to generate pathogenic mutations and harmful bio-agents. Hence, we declare that AlphaDE should be restricted to research purposes, and any applications should undergo comprehensive experiments and human inspections.


\bibliography{example_paper}
\bibliographystyle{icml2026}

\newpage
\appendix
\onecolumn

\section{Descriptions of Benchmark Tasks}
\label{appendix:benchmark_tasks}
Here we briefly describe the eight benchmark tasks constructed by Ren et al. \citep{ren_proximal_2022}.
\begin{itemize}
    \item Green Fluorescent Proteins (avGFP). Green Fluorescent Proteins from \textit{Aequorea victoria}, which can exhibit bright green fluorescence when exposed to light in the blue to the ultraviolet range, are used as biosensors to detect gene expressions and protein locations. Here, we optimize the wild-type sequence in the search space with a size of $20^{238}$ to get higher fluorescence intensity.
    \item Adeno-associated Viruses (AAV). AAVs are a group of small viruses belonging to the family of dependoviruses, which show great potential in the field of gene therapy. Here we optimize a 28-amino acid segment (position 561-588) of the VP1 protein located in the capsid of the Adeno-associated virus to design more capable sequences measured by AAV liabilities.
    \item TEM-1 $\beta$-Lactamase (TEM). TEM-1 $\beta$-Lactamase protein resisting penicillin antibiotics in E.coli is widely studied to understand the mutational effect and fitness landscape \citep{bershtein_robustnessepistasis_2006,jacquier_capturing_2013}. Here we optimize the thermodynamic stability in the protein sequence space with a size of $20^{286}$.
    \item Ubiquitination Factor Ube4b (E4B). Ubiquitination factor Ube4b plays an important role in the trash degradation process in the cell by interacting with ubiquitin and other proteins. We focus on designing E4B with higher enzyme activity. The size of the search space is $20^{102}$.
    \item Aliphatic Amide Hydrolase (AMIE). Amidase encoded by amiE is an industrially-relevant enzyme from \textit{Pseudomonas aeruginosa}. We seek to optimize amidase sequences that lead to great enzyme activities, which defines a search space with $20^{341}$ sequences.
    \item Levoglucosan Kinase (LGK). Levoglucosan kinase converts LGK to the glycolytic intermediate glucose-6-phosphate in an ATP-dependent reaction. Here we optimize in a protein sequence space of $20^{439}$ for improving enzyme activity fitness.
    \item Poly(A)-binding Protein (PAB1). PAB1 functions by binding to multiple adenosine monophosphates (poly-A) using the RNA recognition motif. We optimize to improve binding fitness. The search space size is $20^{75}$ on a segment of the wild-type sequence.
    \item SUMO E2 conjugase (UBE2I). Using human SUMO E2 conjugase to map the functions of human genomes is significant for scientific research. We improve the fitness measured by growth rescue rate at high temperature in a yeast strain with a search space sized $20^{159}$.
\end{itemize}

\section{Details of Homologous Sequence Datasets}
\label{appendix:dataset}
\begin{figure}[htbp]
    \centering
    \includegraphics[width=1.0\linewidth]{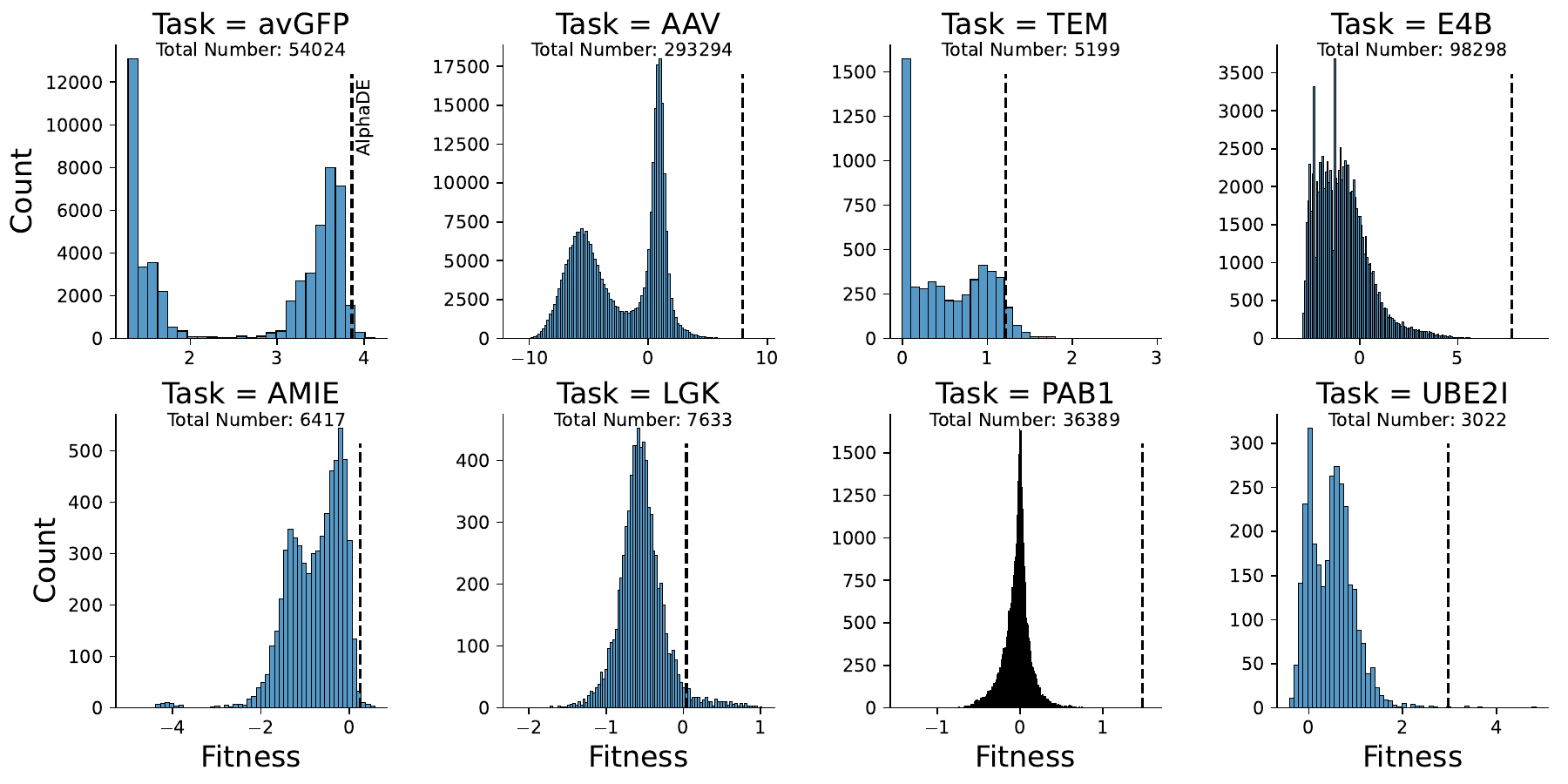}
    \caption{The sequence fitness distributions of each protein dataset used in each task. The black dashed line indicates the fitness location of the best sequences generated by AlphaDE.}
    \label{fig:data_distribution}
\end{figure}

Here, we describe the homologous sequence datasets of the eight benchmark tasks, which are used in the fine-tuning step of AlphaDE. Their fitness distributions and total numbers of sequences are plotted in Figure~\ref{fig:data_distribution}. Please note that we do not use these sequences' fitness labels in either the pretrain, fine-tune, or MCTS inference steps of AlphaDE. We use these protein sequences only for the unsupervised masked language modeling learning in the fine-tuning step. At the same time, we show AlphaDE works with homologous sequences retrieved from the homology search in Appendix~\ref{appendix:phmmer}.

\section{Hyperparameter Settings of AlphaDE}
\label{appendix:hyperparameter}
Here we list the hyperparameters in the fine-tuning step and MCTS inference step of AlphaDE. As there are many hyperparameters in both the protein language model and MCTS, we always try to follow the common practice of hyperparameters in previous works to mitigate the effort to tune hyperparameters. We found the hyperparameter settings of AlphaDE work well across tasks without specific tuning for each task.

\subsection{Hyperparameters of Fine-Tuning}
For the hyperparameters of fine-tuning, we follow the default setting in the fine-tuning script given by the Transformers package (the most widely used Python package for developing large language models). The fine-tuning script using a masked language modeling loss is available on \href{https://github.com/huggingface/transformers/tree/main/examples/pytorch/language-modeling}{GitHub}. We fine-tune the pretrained protein language models on each task's homologous sequence dataset for 3 epochs and the batch size per device is set to 8. The learning rate is $5 \times 10^{-5}$ and the optimizer is AdamW. The learning rate scheduler type to use is linear while the weight decay is not enabled. These hyperparameter settings in the fine-tuning step are the same for the ESM2-series, ProtBert, and ESM-1b models. These fine-tuning hyperparameters (learning rate, optimizer, epochs, and so on) are default configurations in large language models and work well in our setting. We only set the batch size per device to 8 for low GPU memory cost. The GPU memory should be slightly larger than 32GB to ensure the fine-tuning on the LGK task with the longest protein sequence, with the protein language model size of 650M. We do not fine-tune the ESM2-3B model and the ESM2-15B model, as fine-tuning the two models requires GPUs with much larger memory. At the same time, the results from Table~\ref{tab:esm2size} indicate that the small fine-tuned ESM2-35M model owns enough evolution ability to help AlphaDE evolve proteins.

\subsection{Hyperparameters of MCTS}
For the hyperparameters of the MCTS inference step in AlphaDE, we follow the default setting in EvoPlay. The constant $c$ in Eq.~(\ref{eq:puct}) is a trade-off coefficient to balance exploitation and exploration and is set to 10 for all tasks. The number of simulation rollout times in the Rollout step is set to 200. For an episode, there are several termination conditions. First, AlphaDE terminates when it meets the maximum tree depth (i.e., maximum mutation number), which is set at 100. Second, if the current move is invalid when the state sequence remains unchanged or changes to a previously generated sequence, the current episode terminates. The third termination condition is that the current mutated sequence's fitness value is smaller than the fitness value of the sequence it mutates from. We also test the sensitivity of key hyperparameters of MCTS, including exploration constant, tree depth, and rollout number, in Appendix~\ref{appendix:hyperparameter_sensitivity}.

For the training of the value network, the update is performed after each episode. The number of training steps for each update is 5, while the batch size for each training step is set to 32. The size of the replay buffer while stores experiences is set to 10000. Here, the optimizer is Adam and the learning rate is $2 \times 10^{-3}$. The weight decay is $1 \times 10^{-4}$. The loss is calculated based on the mean squared error between the predicted fitness value by the value network and the simulated fitness value by the oracle model. These hyperparameter settings in the MCTS inference step are the same for the AlphaDE with ESM2-series, ProtBert, and ESM-1b models. For the MCTS inference step, the GPU memory of less than 8 GB is enough for all tasks.

\subsection{Value Network Settings in AlphaDE}
The value network of AlphaDE is with the same network architecture as in EvoPlay \citep{wang_self-play_2023}, which consists of 4 convolutional neural network layers and 2 dense layers. As the same network architecture works well in AlphaDE, we do not modify the value network architecture. The only difference is that we do not specify the activation function in the output layer, as we found that the predicted fitness value of different tasks has different value ranges, and we do not restrict the output range of the value network.

\section{AlphaDE with Different EMS2 Model Sizes}
\label{appendix:esm2size}

We also provide the performance of AlphaDE with other ESM2-series models of different sizes such as 8M, 150M, and 650M in Table~\ref{tab:esm2size}. We find no significant differences between different ESM2 versions while ESM2-35M performs stably, indicating that the evolution information encoded in the fine-tuned ESM2-35M is enough for AlphaDE to conduct an effective tree search for directed evolution in tested protein tasks.

\begin{table}[htbp]
\centering
\caption{Performance of AlphaDE with different ESM2 model sizes. We present the maximum fitness scores obtained in 1000 black-box oracle queries. Results are averaged over five independent trials.}
\begin{tabular}{lllllllll}
\hline
AlphaDE Model & avGFP & AAV & TEM & E4B & AMIE & LGK & PAB1 & UBE2I \\ \hline
ESM2-8M & 3.86 & 8.30 & 1.21 & 7.86 & 0.24 & 0.02 & 1.14 & 2.90 \\ \hline
ESM2-35M & 3.86 & 7.95 & 1.22 & 7.75 & 0.24 & 0.04 & 1.47 & 2.97 \\ \hline
ESM2-150M & 3.86 & 7.96 & 1.22 & 7.68 & 0.25 & 0.04 & 1.64 & 2.89 \\ \hline
ESM2-650M & 3.86 & 8.15 & 1.22 & 7.67 & 0.25 & 0.04 & 1.08 & 2.84 \\ \hline
\end{tabular}
\label{tab:esm2size}
\end{table}

\section{Fitness Oracles}
The use of TAPE and ESM-1b oracles follows the standard evaluation procedure in previous works \citep{ren_proximal_2022,wang_self-play_2023,qiu_tree_2024}. The ESM-1b oracle is used in the avGFP condensing experiment as it supports some special characters, including the token `-'. We regard the character token `-' as an amino acid residue deletion and utilize ESM-1b to predict the fitness of the condensed avGFP sequences. These oracle models and their download script can be obtained from \href{https://github.com/HeliXonProtein/proximal-exploration}{GitHub}.

\section{Ablation Study of AlphaDE}
\label{appendix:ablation}

In this section, we conduct the ablation study to validate each component of AlphaDE. Specifically, EvoPlay is an advanced MCTS framework for directed evolution without the protein language model as mutation guidance, where an actor network is learned from scratch. Therefore, to quantify the contribution of the pretrained protein language model for mutation action guidance, we compare AlphaDE (MCTS that uses pretrained ESM2) in Section~\ref{section:zeroshot_scaling} with EvoPlay (MCTS that uses an actor from scratch). We see AlphaDE with pretrained ESM2 already outperforms EvoPlay in most tasks. To quantify the contribution of the fine-tuned protein language model, we compare the results of AlphaDE with fine-tuned ESM2 in Table~\ref{tab:benchmark} and AlphaDE with pretrained ESM2 in Figure~\ref{fig:pretrained_scaling_law}. We extract these results to include in Table~\ref{tab:ablation_finetuning} below for a direct comparison. It is clear that the fine-tuning step in AlphaDE contributes much to the performance.

\begin{table}[htbp]
\centering
\caption{Ablation study of the fine-tuning step in AlphaDE. We present the maximum fitness scores obtained in 1000 black-box oracle queries. Results are averaged over five independent trials.}
\begin{tabular}{lllllllll}
\hline
\begin{tabular}[c]{@{}l@{}}Ablation of fine-tuning step \end{tabular} & avGFP & AAV & TEM & E4B & AMIE & LGK & PAB1 & UBE2I \\ \hline
\begin{tabular}[c]{@{}l@{}}MCTS with fine-tuned \\ ESM2-35M model (AlphaDE)\end{tabular} & \textbf{3.86} & \textbf{7.95} & \textbf{1.22} & \textbf{7.75} & \textbf{0.24} & \textbf{0.04} & \textbf{1.47} & \textbf{2.97} \\ \hline
\begin{tabular}[c]{@{}l@{}}MCTS with pretrained \\ ESM2-35M model (AlphaDE)\end{tabular} & 3.13 & -0.72 & 0.89 & 0.38 & -7.08 & -0.01 & 0.82 & 1.70 \\ \hline
\begin{tabular}[c]{@{}l@{}}MCTS with actor network \\ learning from scratch (EvoPlay)\end{tabular} & 1.72 & -3.45 & 0.01 & -0.40 & -0.88 & -1.09 & 0.34 & 1.87 \\ \hline
\end{tabular}
\label{tab:ablation_finetuning}
\end{table}

At the same time, we also conduct the ablation of the whole MCTS search part with greedy search and beam search. The results are shown in Table~\ref{tab:ablation_mcts}, which highlights the contribution of the MCTS search in AlphaDE.

\begin{table}[htbp]
\centering
\caption{Ablation study of the MCTS inference step in AlphaDE. We present the maximum fitness scores obtained in 1000 black-box oracle queries. Results are averaged over five independent trials.}
\begin{tabular}{lllll}
\hline
Ablation of MCTS & avGFP & UBE2I & E4B & PAB1 \\ \hline
\begin{tabular}[c]{@{}l@{}}AlphaDE (MCTS with fine-tuned ESM2-35M model)\end{tabular} & \textbf{3.865} & \textbf{2.975} & \textbf{7.746} & \textbf{1.469} \\ \hline
\begin{tabular}[c]{@{}l@{}}Greedy search with fine-tuned ESM2-35M model\end{tabular} & 3.863 & 2.967 & 1.112 & 0.216 \\ \hline
\begin{tabular}[c]{@{}l@{}}Beam search with fine-tuned ESM2-35M model\end{tabular} & 3.813 & 1.310 & 5.247 & 0.207 \\ \hline
\end{tabular}
\label{tab:ablation_mcts}
\end{table}

\section{Benchmark Experiments with ESM-1b Oracle}
\label{appendix:esm1b_oracle}

We conduct the benchmark experiments with ESM-1b oracle taken from \citet{ren_proximal_2022}. Results are shown in Table~\ref{tab:esm1b_oracle}. AlphaDE again beats various competitive baselines in the benchmark setting of EMS-1b oracle.

\begin{table}[htbp]
\centering
\caption{Benchmark experiments with another ESM-1b oracle. We present the maximum fitness scores obtained in 1000 black-box oracle queries. Results are averaged over five independent trials.}
\begin{tabular}{llllllll}
\hline
Task & AlphaDE & LatentDE & MLDE & TreeNeuralTS & TreeNeuralUCB & EvoPlay & PEX \\ \hline
avGFP & \textbf{3.991} & 3.677 & 2.443 & 3.707 & 3.785 & 3.835 & 3.754 \\ \hline
UBE2I & \textbf{3.989} & 2.360 & 1.434 & 3.632 & 3.653 & 0.647 & 3.101 \\ \hline
\end{tabular}
\label{tab:esm1b_oracle}
\end{table}

\section{Hyperparameter Sensitivity}
\label{appendix:hyperparameter_sensitivity}
We study the sensitivity of several important hyperparameters in AlphaDE's MCTS part, including the exploration constant, tree depth, and rollout number. Results are averaged over five independent trials. We use the top 1000 sequences because the results of the top 1 are good enough to be less discriminative.

\subsection{Hyperparameter Sensitivity of Exploration Constant}

In Equation~\ref{eq:puct}, the constant $c$ balances the exploration and exploitation. Here we conduct a hyperparameter sensitivity study to investigate how $c$ influences the performance of AlphaDE. We run AlphaDE with different $c$ values such as 0.1, 1, 10, 100, and 1000 in the avGFP and UBE2I tasks. The results are summarized in Table~\ref{tab:sensitivity_c}. We see that if $c$ is too large such as 100 and 1000, AlphaDE's performance decreases significantly. Otherwise, AlphaDE achieves similar performance when $c=$ 0.1, 1, and 10. In this paper, we set $c=$10 as a default value and do not specifically tune $c$ for each task. However, we should consider that the best $c$ value may differ in tasks and require further investigation when running AlphaDE on a specific task.

\begin{table}[htbp]
\centering
\caption{Sensitivity study of the exploration constant $c$. We present the average fitness scores obtained in 1000 black-box oracle queries. Results are averaged over five independent trials.}
\begin{tabular}{llllll}
\hline
$c$ & 0.1 & 1 & 10 (default) & 100 & 1000 \\ \hline
avGFP & 3.72 & 3.73 & 3.72 & 3.59 & 3.49 \\ \hline
UBE2I & 2.60 & 2.45 & 2.49 & 2.03 & 1.06 \\ \hline
\end{tabular}
\label{tab:sensitivity_c}
\end{table}

\subsection{Hyperparameter Sensitivity of Tree Depth}

For the tree depth, we give the sensitivity study results in Table~\ref{tab:sensitivity_tree_depth}. We see that the tree depth does not affect AlphaDE’s performance much, which indicates that AlphaDE is robust to the tree depth.

\begin{table}[htbp]
\centering
\caption{Sensitivity study of the tree depth. We present the average fitness scores obtained in 1000 black-box oracle queries. Results are averaged over five independent trials.}
\begin{tabular}{llllll}
\hline
tree depth & 1 & 5 & 10 & 100 (default) & 1000 \\ \hline
avGFP & 3.71 & 3.73 & 3.70 & 3.72 & 3.70 \\ \hline
UBE2I & 2.46 & 2.51 & 2.45 & 2.49 & 2.67 \\ \hline
\end{tabular}
\label{tab:sensitivity_tree_depth}
\end{table}

\subsection{Hyperparameter Sensitivity of Rollout Number}

For the rollout number, we give the sensitivity study results in Table~\ref{tab:sensitivity_rollout_number}. We see that the rollout number does not affect AlphaDE’s performance much, which indicates that AlphaDE is robust to the rollout number. Meanwhile, we also notice that at the extreme setting where the rollout number is 1 in the task of avGFP, the performance of AlphaDE decreases significantly, which necessitates the importance of multiple rollouts at the reached leaf node.

\begin{table}[htbp]
\centering
\caption{Sensitivity study of the rollout number. We present the average fitness scores obtained in 1000 black-box oracle queries. Results are averaged over five independent trials.}
\begin{tabular}{lllllll}
\hline
rollout number & 1 & 10 & 50 & 100 & 200 (default) & 1000 \\ \hline
avGFP & 2.26 & 3.70 & 3.70 & 3.72 & 3.72 & 3.71 \\ \hline
UBE2I & 2.43 & 2.42 & 2.35 & 2.49 & 2.49 & 2.41 \\ \hline
\end{tabular}
\label{tab:sensitivity_rollout_number}
\end{table}

\section{Diversity of Evolved Sequences by AlphaDE}
\label{appendix:seq_diversity}
Here, we give the diversity of the evolved sequences by AlphaDE. The diversity is calculated by the top $K$ sequences from each trial and there are 5 trials for each task. The top sequences are ranked by the fitness value. Then diversity equals to the number of unique sequences divided by $5 \times K$. 100\% indicates all the generated sequences are different. $K$ is set at 1, 10, 100, and 1000 respectively and the results are shown in Table~\ref{tab:seq_diversity}. We see that AlphaDE generates diverse protein sequences while the top 1 sequences from each trial are different.

\begin{table}[htbp]
\centering
\caption{Diversity of generated sequences by AlphaDE. The diversity is calculated with sequences collected from 5 trials. The top sequences are ranked according to their fitness values.}
\begin{tabular}{lllllllll}
\hline
Top & avGFP & AAV & TEM & E4B & AMIE & LGK & PAB1 & UBE2I \\ \hline
1 & 100.00\% & 100.00\% & 100.00\% & 100.00\% & 100.00\% & 100.00\% & 100.00\% & 100.00\% \\
10 & 100.00\% & 100.00\% & 100.00\% & 100.00\% & 100.00\% & 100.00\% & 100.00\% & 100.00\% \\
100 & 99.60\% & 96.80\% & 100.00\% & 100.00\% & 100.00\% & 100.00\% & 100.00\% & 100.00\% \\
1000 & 96.96\% & 93.80\% & 99.14\% & 98.96\% & 99.12\% & 93.22\% & 98.82\% & 99.64\% \\ \hline
\end{tabular}
\label{tab:seq_diversity}
\end{table}

On the other hand, we are also interested in the repetition rate that how often the generated sequences of AlphaDE exist in each task's fine-tuning dataset. The repetition rate is calculated by the top $K$ sequences from each trial, and there are 5 trials for each task. The top sequences are ranked by the fitness value. Then the repetition rate equals to the number of repeated sequences between the top $K$ sequences and the fine-tuning dataset divided by $5 \times K$. $K$ is set at 1, 10, 100, and 1000 respectively, and the results are shown in Table~\ref{tab:seq_repetition}. From Table~\ref{tab:seq_repetition}, we see that, in most tasks, the repetition rate maintains at a very low level. This indicates that AlphaDE generates novel sequences instead of repeating the sequences from the fine-tuning dataset. We also note that, as an exception, for the top 1 sequences of task AAV, the repetition rate is relatively high. But this high repetition rate drops as $K$ increases. For example, the repetition rate of the top 10 AAV sequences drops to 26\%, which means the high-fitness sequences are mostly different from the fine-tuning sequences.

\begin{table}[htbp]
\centering
\caption{Repetition rate of generated sequences by AlphaDE. The repetition rate is calculated with sequences collected from 5 trials. The top sequences are ranked according to their fitness values.}
\begin{tabular}{lllllllll}
\hline
Top & avGFP & AAV & TEM & E4B & AMIE & LGK & PAB1 & UBE2I \\ \hline
1 & 0.00\% & 80.00\% & 0.00\% & 0.00\% & 0.00\% & 0.00\% & 0.00\% & 0.00\% \\
10 & 0.00\% & 26.00\% & 0.00\% & 2.00\% & 0.00\% & 0.00\% & 0.00\% & 0.00\% \\
100 & 0.00\% & 8.60\% & 1.20\% & 0.40\% & 0.00\% & 0.00\% & 0.00\% & 0.00\% \\
1000 & 0.62\% & 7.64\% & 3.80\% & 3.52\% & 0.82\% & 0.00\% & 1.16\% & 1.66\% \\ \hline
\end{tabular}
\label{tab:seq_repetition}
\end{table}

\section{Fine-tuning with Sequences from Homology Searching}
\label{appendix:phmmer}

\begin{table}[htbp]
\centering
\caption{Results of different fine-tuning datasets. We present the maximum fitness scores obtained in 1000 black-box oracle queries. AlphaDE (phmmer) is also averaged over five independent trials.}
\begin{tabular}{llllll}
\hline
Method & AlphaDE (DMS) & AlphaDE (phmmer) & PEX (CNN) & AdaLead & TreeNeuralTS \\ \hline
avGFP & 3.86 & 3.83 & 3.30 & 1.81 & 2.44 \\ \hline
\end{tabular}
\label{tab:phmmer}
\end{table}

In the benchmark experiments, almost all the fine-tuned sequence datasets are from deep mutational scanning (DMS), which are not always available for most proteins. Therefore, in this section, we introduce the homology searching technique to construct the dataset of homologous sequences for the fine-tuning step in AlphaDE. Next, we use avGFP as an example. Specifically, we use the phmmer homology searching procedure of the biosequence analysis tool HMMER \citep{potter_hmmer_2018} to find homologous sequences of the starting weakest sequence in the avGFP task. We use the default settings of phmmer to search the SwissProt database, UniProt database, Reference Proteomes database, and PDB database. After filtering sequences with the same length as avGFP and removing three duplicate sequences in the DMS dataset, we found 236 unique sequences to construct the avGFP phmmer dataset, which has no overlap with the avGFP DMS dataset. Then we follow the standard AlphaDE fine-tuning step on the avGFP phmmer dataset and perform the MCTS step with the fine-tuned model (here $c$ is set at 1.0 for a better performance). Results are shown in Table~\ref{tab:phmmer} below. We see that, AlphaDE, which uses the phmmer homologous sequence dataset for fine-tuning, also achieves superior performance.

\section{Details of Condensing avGFP}
\label{appendix:avgfp}

When initializing the deleted avGFP sequences, we keep the $\beta$-barrel residues and the chromophore-related residues. The $\beta$-barrel residues involve the residue 1 to residue 38. The key chromophore residues are residue 65, 66, and 67 \citep{hayes_simulating_2025}, and we set the chromophore-related residues to be residue 58 to residue 74. Then we delete half of the remaining residues, which results in a starting deleted sequence with length 146. In contrast, the wild-type sequence has a length of 238 residues. The wild-type avGFP sequence, the starting deleted sequence, and the final filtered ccGFP1-5 sequences are given in Table~\ref{tab:avgfp_seqs}. When filtering, the amino acid sequences of ccGFP1-5 are fed into \href{https://alphafoldserver.com/}{AlphaFold 3 server} to predict their structures. Then we align these structures with the wild-type structure (PDB ID: \href{https://www.rcsb.org/structure/1ema}{1EMA}) by PyMol to calculate the RMSD distances. Although this is a computational proof-of-concept task, it shows the great potential of AlphaDE for different purposes with directed evolution.

\begin{table}[htbp]
\centering
\caption{Amino acid sequences of avGFP variants in the protein sequence condensing experiment.}
\begin{tabular}{|p{1.5cm}|p{11.5cm}|}
\hline
Name & Amino Acid Sequence \\ \hline
wild-type & {\tiny\seqsplit{MSKGEELFTGVVPILVELDGDVNGHKFSVSGEGEGDATYGKLTLKFICTTGKLPVPWPTLVTTLSYGVQCFSRYPDHMKQHDFFKSAMPEGYVQERTIFFKDDGNYKTRAEVKFEGDTLVNRIELKGIDFKEDGNILGHKLEYNYNSHNVYIMADKQKNGIKVNFKIRHNIEDGSVQLADHYQQNTPIGDGPVLLPDNHYLSTQSALSKDPNEKRDHMVLLEFVTAAGITHGMDELYK}} \\ \hline
starting & {\tiny\seqsplit{MSKGEELFTGVVPILVELDGDVNGHKFSVSGEGEGDATGLLFCTKPPPTLVTTLSYGVQCFSRYDMQDFSMEYQRIFDGYTAVFGTVREKIFEGIGKENNHVIAKKGKNKRNEGVLDYQTIDPLPNYSQASDNKDMLEVAGTGDLK}} \\ \hline
ccGFP1 & {\tiny\seqsplit{MSKGEELFTLVVPILVELRGDVNGHKFSVSGEGEGNATGLTLKFCTTGKLPVPWPTLVTTLSYGVQCFSRYDVMQHDFKSAMEGYVQRTIFFDGYTRAEVFGDTVRELKGIFEGIGKENNSHNVWIADKKGIKNFKRNEGSVVADHYQTFIDPVLPNILSTQSASDNKRDHMILLEGVAGHHGMDLYK}} \\ \hline
ccGFP2 & {\tiny\seqsplit{MSKGEELFTLVVPILVELRGDVNGHKFSVTGEGEGNATGLTLKFCTTGKLPVPWPTLVTTLSYGVQCFSRYDVMQHDFKSAMEGYVQRTIFFDGYTRAEVFGDTVRELKGIFEGIGKENNSHNVWIADKKGIKNFKRNEGSVVADHYQTFIDPVLPNILSTQSASDNKRDHMILLEGVAGHHGMDLYK}} \\ \hline
ccGFP3 & {\tiny\seqsplit{MSKGEELFTGVVPILVELDGDVNGHKFSVSGEGEGDATGQLTLKFCTTKLPVAWPTLVTTPSYGVQCFSRYDMKQHDQSAMEYAQRDIFFKDYTRAVKFGTLVRELKVIDFEGNILGKEYNNSHVIADKQKGIKNKRNEGVQLDHYQQNTPIVDPVLLPLNHYSQSALSDNEKRDMLLEFVTAGTGLK}} \\ \hline
ccGFP4 & {\tiny\seqsplit{MSKGIELFTGVPILVELDGDVNGHKFSVSGEGEGDASGKLLFCTTKPVPCTLVTTLSYGVQCFSRYPDMKQHDFKSAMRYQRRTIFDGNYTAVFGTLVRIELKGIDFKEGIGKENYNSHVIMADKQKNGIKVNFKRHTIEGVLADHYQTIDGPVLPNHYLSTQASIDNKDMVLEVTAAGTHGMDLK}} \\ \hline
ccGFP5 & {\tiny\seqsplit{MSKGIELFTGVPILVELDGDVNGHKFSVSGEGEGDASGKLLFCTTKPVPCTLVTTLSYGVQCFSRYPDMKQHDFKSAMRYQRRTIFDGNYTAVFGTLVRIELKGIDFKEGIGKENYNSHVIMADKQKNGKVNFKRHTIEGVLADHYQTIDGPVLPNHYLSTQASIDNKDMVLEVTAAGTHGMDLK}} \\ \hline
\end{tabular}
\label{tab:avgfp_seqs}
\end{table}

At the same time, we also compare with AlphaDE with the pretrained protein language model to validate whether it recovers the folded structure of wild-type avGFP. We follow the same pipeline as condensing avGFP with standard AlphaDE. The comparison of the AlphaDE with pretrained ESM2-35M and standard AlphaDE is shown in Table~\ref{tab:condense_comparison} below. It is clear that the condensed avGFPs by AlphaDE with pretrained ESM2-35M have much larger RMSDs, indicating the recovered folded structures are less similar to the wild-type avGFP structure than the standard AlphaDE.

\begin{table}[htbp]
\centering
\caption{Comparison of AlphaDE with pretrained ESM2-35M and with fine-tuned ESM2-35M for condensing avGFP. The metric is RMSD to the wild-type avGFP (PDB ID: 1EMA).}
\begin{tabular}{llllll}
\hline
RMSD ($\downarrow$) & 1 & 2 & 3 & 4 & 5 \\ \hline
standard AlphaDE with fine-tuned ESM2-35M & 1.10 & 1.40 & 1.70 & 1.83 & 1.92 \\ \hline
AlphaDE with pretrained ESM2-35M & 5.92 & 6.27 & 6.79 & 7.55 & 8.08 \\ \hline
\end{tabular}
\label{tab:condense_comparison}
\end{table}

\section{Computational Efficacy of AlphaDE}
\label{appendix:efficacy}

Here, we give the running time of the fine-tuning step and the MCTS inference step. The protein language model here is ESM2-35M, which is the default configuration in AlphaDE. The running time is shown in Table~\ref{tab:running_time}. For the fine-tuning step, the running time mainly depends on the number of sequences in the dataset. We use three NVIDIA GPUs for this fine-tuning step and one NVIDIA GPU for the MCTS inference step. When fine-tuning, the GPU memory cost depends on the size of protein language models and the length of protein amino acid sequences.

\begin{table}[htbp]
\centering
\caption{Running time of AlphaDE with ESM2-35M. The unit of running time is the hour.}
\begin{tabular}{lllllllll}
\hline
Step & avGFP & AAV & TEM & E4B & AMIE & LGK & PAB1 & UBE2I \\ \hline
Fine-Tuning & 0.38 & 1.22 & 0.04 & 0.45 & 0.06 & 0.10 & 0.17 & 0.02 \\
MCTS Inference & 4.36 & 1.66 & 4.46 & 3.04 & 3.93 & 1.85 & 1.12 & 1.08 \\ \hline
\end{tabular}
\label{tab:running_time}
\end{table}

Meanwhile, we additionally provide the computational cost of competitive baselines such as TreeNeuralTS and TreeNeuralUCB for a comparison in the task avGFP. We also compare with EvoPlay. The results are included in Table~\ref{tab:running_time_baselines}. As AlphaDE utilizes the protein language model, it is expected to take a longer time to run. We also see that the running time of AlphaDE is acceptable, compared with other baselines.

\begin{table}[htbp]
\centering
\caption{Running time of AlphaDE and different baselines in avGFP averaged over 5 trials. The unit is the hour.}
\begin{tabular}{lllll}
\hline
Task & AlphaDE & TreeNeuralTS & TreeNeuralUCB & EvoPlay \\ \hline
avGFP & 4.74 & 2.33 & 1.79 & 0.69 \\ \hline
\end{tabular}
\label{tab:running_time_baselines}
\end{table}

\section{Limitation}
\label{appendix:limitation}
Our study has limitations under extensive consideration. First, the oracle models for fitness evaluation may have biases and cannot replace the real-world wet-experiment measurements. Second, the fine-tuning step requires homologous sequences, which may not always exist for a specific protein. If there are novel or poorly characterized proteins with seldom homologous sequences, the application of our method may be restricted. Luckily, AlphaDE supports the few-shot fine-tuning as indicated in Section~\ref{section:fewshot_finetuning}, which greatly reduces the needed number of homologous sequences. Additionally, as in Section~\ref{section:zeroshot_scaling}, our method supports zero-shot mode with the pretrained protein language models if homologous sequences are not available.


\end{document}